\documentclass[runningheads]{llncs}
\usepackage{graphicx}
\usepackage[table]{xcolor}
\usepackage{newpxtext}
\usepackage{newpxmath}
\usepackage[export]{adjustbox}
\usepackage{setspace}
\usepackage{pdflscape}
\usepackage{afterpage}
\usepackage{pbox}
\usepackage{pythonhighlight}
\usepackage{algpseudocode}
\usepackage{algorithm}
\usepackage{hhline}
\usepackage{multirow}
\usepackage{booktabs}
\usepackage{amsfonts}
\usepackage{todonotes}
\usepackage[skins]{tcolorbox}
\usepackage{csquotes}
\usepackage[symbols,nogroupskip,sort=none]{glossaries-extra}
\usepackage{multicol}
\usepackage{biblatex} %Imports biblatex package
\begin{document}
\title{Unveiling the Potential of Counterfactuals Explanations in Employability}
\author{Raphael Mazzine Barbosa de Oliveira \and
Sofie Goethals \and
Dieter Brughmans\and
David Martens
}
\authorrunning{Raphael Mazzine et al.}
\institute{Department of Engineering Management, University of Antwerp, Prinsstraat 13, Antwerp, 2000, Belgium \\
\email{Raphael.MazzineBarbosaDeOliveira@uantwerpen.be}}
\maketitle              % typeset the header of the contribution
\begin{abstract}
In eXplainable Artificial Intelligence (XAI), counterfactual explanations are known to give simple, short, and comprehensible justifications for complex model decisions. However, we are yet to see more applied studies in which they are applied in real-world cases. To fill this gap, this study focuses on showing how counterfactuals are applied to employability-related problems which involve complex machine learning algorithms. For these use cases, we use real data obtained from a public Belgian employment institution (VDAB). The use cases presented go beyond the mere application of counterfactuals as explanations, showing how they can enhance decision support, comply with legal requirements, guide controlled changes, and analyze novel insights.

\end{abstract}
\section{Introduction}
Explainable artificial intelligence has applications as long as machine learning methods are part of a problem. Therefore, given the widespread popularity of models like neural networks and random forests, it is not difficult to find use cases. In this article, we will talk about the employability field, which has been defined as "an individual's ability to obtain and maintain initial employment, move between roles within the same organization, obtain new employment if necessary, and/or generally secure suitable and satisfying work"~\cite{hillage1998employability}. 

Both job candidates and potential employers enter the job market with differing knowledge, competence, and abilities, collectively referred to as "skills"~\cite{pheko2017addressing,klosters2014matching}. As a result of the imbalance between supply and demand in all economies~\cite{pheko2017addressing,klosters2014matching}, it is challenging to match job applicants with available positions. Current trends indicate an increasing shift towards the demand-driven provision of education, where employers also have a voice, resulting in the participation of multiple stakeholders (with different interests and objectives). 

Due to the complexity of these issues, Artificial Intelligence (AI) is regarded as a technology that can aid in the resolution of numerous employability-related obstacles. Reported solutions attempt to leverage AI to this end: predictive methods are used to estimate whether an individual with particular skills would meet market demands~\cite{jantawan2013application}; machine-learning-based classifiers are used to label job advertisements~\cite{boselli2018classifying}; intelligent chatbot systems are developed for HR service delivery~\cite{mujtaba2019ethical}; and AI-based recommender systems~\cite{martinez2018recommendation} work by pre-filtering a candidate pool~\cite{ochmann2021evaluation}. These examples demonstrate AI's growing significance in research and industry. However, their "black box" nature makes it extremely difficult, if not impossible, to explain their predictions~\cite{goethals2022comprehensibility}. This opaqueness can be crucial for regulatory affairs~\cite{wachter2017counterfactual} and has hindered the discovery of unfair algorithmic biases that can be present in models that are biased against certain groups, which could lead to widespread discrimination based on race or gender in hiring practices~\cite{rai2020explainable}.

Explainable Artificial Intelligence (XAI) has emerged to investigate potential solutions for these deficiencies, and multiple methodologies~\cite{lundberg2017unified,ribeiro2016should} have been developed in recent years. This paper focuses on a popular local, model-agnostic post-hoc method, namely counterfactual explanations. A counterfactual explanation can be defined as the minimal and irreducible set of input features that must be altered to change the predicted class of an observation~\cite{martens2014explaining}. For instance, a person denied a job could provide a counterfactual explanation: \textit{If your skills included Python and Cloud Computing, the prediction would change from unsuitable to suitable for this data science position}.

Given its simplicity, this type of explanation is well-suited for how humans prefer to receive explanations~\cite{miller2019explanation} and satisfies the GDPR requirement for providing explanations for automated decision-making systems~\cite{wachter2017counterfactual}. In addition, counterfactual explanations can serve purposes other than elucidating predictions.
For example, we can use counterfactuals to guide a feasible path to alter output decisions~\cite{karimi2020algorithmic} and as a bias detection tool~\cite{sokol2019counterfactual}. Taking into account its simplicity and adaptability, we identify several applicable use cases for counterfactual explanations involving diverse requirements, methodological implications, and relevant stakeholders. To illustrate how they can be applied in real-world scenarios, we use a dataset containing over 12 million job listings from a Belgian employment agency to examine several use cases. We demonstrate that the resulting explanations are promptly provided, which is essential for real-world applications, and that they are sparse compared to other popular XAI methodologies (LIME and SHAP). This property is frequently claimed to be advantageous for producing understandable explanations. Therefore, we emphasize that this study exemplifies novel applications of counterfactual explanations in the field of employability that extend beyond their use as a decision explanation methodology. Moreover, for each use case, we further expand the discussion outside the employability field, generalizing the applicability of counterfactual explanations to any relevant problem.

\section{Materials and Methods}
VDAB (Vlaamse Dienst voor Arbeidsbemiddeling en Beroepsopleiding), a public employment service in Flanders, Belgium, provided the raw datasets used to construct the model and generate counterfactual instances. This source is proprietary (secondary) and comprises pseudorandomized HR data that contain detailed information about the skills profiles of job seekers and vacancies~\cite{mashayekhi2021quantifying}. As the dataset is proprietary and subject to terms and conditions, such as a Data Processing Agreement, its processing and publication have legal limitations to its usage and publication. The processed datasets include job requirements as binary column features, where 0 indicates that the position does not require the skill and 1 indicates that it does. Our dataset label is a number representing the reach of jobs that a particular set of skills has, i.e., the number of jobs partially or entirely fulfilled by the skills present in the row (equal to 1). 

The dataset contains 1,396,179 rows, where each column represents a skill, with an average of 11.04 mentioned skills per position. The skills characteristics in the dataset are divided into four main categories: studies, study areas, competencies, and languages. The studies include all academic degrees recognized in Belgium, from high school diplomas to bachelor's, master's, and doctoral degrees. The study area is an aggregation level to multiple studies; for example, in the IT study area, bachelor's degrees in software engineering and master's degrees in computer science are grouped together. The competencies include more specific skills, such as mastery of specific software applications, and soft skills, such as team leadership. Finally, it includes information about known languages. The complete dataset (considering all study areas) has 5000 features after preprocessing, of which 4450 are competencies, 500 studies, and 50 languages.

For machine learning tasks, we employ CatBoost, a cutting-edge gradient boosting algorithm for machine learning that is effective for regression problems. The models are constructed using two sets of data: (1) all jobs and (2) only jobs requiring an information technology degree. For each dataset, a regression model was developed utilizing a train/test split of 3/1 and a 3-fold cross-validation grid search for parameter tuning. The test RMSE for the all jobs model was 1,899, while the test RMSE for the information technology jobs model was 249. Since we employ a regression model for a classification problem, we specify a threshold that determines whether or not the instance (set of skills) has a high job reach. In our experiments, high job reach is defined as a number greater than the 90th percentile for the group under consideration. This number is 29,559 for all jobs and 164,150 for IT-related positions.

This paper uses three XAI methodologies: LIME, SHAP, and SEDC. LIME, a feature importance method, is implemented using its original Python package~\cite{ribeiro2016should,ribeiro2016model}, where we describe unfavorable instances for the top 20 most important features while all other parameters are set to their default values. SHAP is another popular feature importance method that is also implemented using its original Python implementation~\cite{lundberg2020local}. We generate the same explanations for SHAP as for LIME, with all other parameters set to their default values. For the counterfactual explanation method, we choose to implement the logic of SEDC, a greedy best-first search algorithm~\cite{martens2014explaining,ramon2020comparison}. We chose SEDC because it is ideally suited to manage a large number of binary features~\cite{martens2014explaining,ramon2020comparison}. Since SEDC is designed for binary classification models, we wrap regression models in which a threshold is used to define a binary class. This approach, where we define a threshold or target in regression models, is frequently used for counterfactual explanations applied to regression models~\cite{spooner2021counterfactual,hada2021exploring}. 

\section{Quantitative Results}
We conducted multiple experiments with distinct data subsets to demonstrate the potential for counterfactual explanations. All experiments are performed on a 2.60GHz Intel(R) Core(TM) i7-6700HQ processor with 16GB of memory and an NVIDIA GeForce GTX 1070 Mobile. In the first experiment, we examine the quantitative metrics of our counterfactual algorithm, namely the speed and sparsity of the resulting explanations, using a sample of 1,000 CVs randomly selected from the test dataset. Taking into account all counterfactuals generated, the average number of changes is 3.33, which is a significantly low value compared to the average number of skills (11.04) and standard deviation (6.7). In terms of time, the average generation time of a counterfactual is 56 seconds (with a standard deviation of 17.5 seconds). The maximum time required to generate a counterfactual is 89 seconds, which is still acceptable given that further scaling improvements can be easily achieved by deploying more high-performance computing equipment.

\begin{table}[ht]
\centering
\begin{tabular}{clll}
\hline
\textbf{Feature changes} & \multicolumn{1}{c}{\textbf{LIME}}  & \multicolumn{1}{c}{\textbf{SHAP}} & \multicolumn{1}{c}{\textbf{SEDC}} \\ \hline
1                        & 3.7 \%                             & 6.9 \%                             & \textbf{9.1 \%}                    \\
2                        & 6.1 \%                             & 14.0 \%                            & \textbf{20.5 \%}                    \\
3                        & 6.3 \%                             & 20.4 \%                            & \textbf{40.2 \%}                    \\
4                        & 6.6 \%                             & 25.0 \%                            & \textbf{97.6 \%}                    \\
5                        & 6.7 \%                             & 27.2 \%                             & \textbf{100 \%}              \\ \hline
\end{tabular}
\caption{Percentage of factual instances where the classification output changes to a favorable outcome when (sequentially) changing up to 5 features indicated by LIME, SHAP, and SEDC.}
\label{tab:vdab_localflipshare}
\end{table}

In Table~\ref{tab:vdab_localflipshare}, we contrast the counterfactual explanations produced by SEDC with the Feature Importance ranking produced by LIME and SHAP. In this comparison, we calculate the proportion of instances in which a previously unfavorable decision was reversed following the modification of the five most relevant (or significant) suggested features. This result, unsurprising given our previous experiments~\cite{martens2014explaining,ramon2020comparison,fernandez2020explaining}, demonstrates the superior performance of counterfactual explanations in locating the smallest subset of relevant features for a class change. In the employment field, where acquiring new skills is generally viewed as costly and time-consuming, these findings provide further support for the superiority of counterfactual explanations over FI explanations.

\section{Counterfactual Use Cases}
In the following sections, we will present multiple use cases where counterfactual explanations can represent a solution or tool for diverse problems in the field of employability. We use this field as a practical example for applications using the real-world data provided by VDAB. We then further expand the discussion for each use case, allowing usage in more general contexts.

\subsection{Counterfactual as Explanations} \label{subsec:exp}
Automated suggestions in consumer services have increased with the development of big data and artificial intelligence.
However, AI-based recommendations can also have unfavorable effects and compromise user autonomy~\cite{alfano2020technologically}. Consequently, measures designed to increase the transparency of automated suggestion decisions are viewed as necessary enhancements. For the employment field, recommendations can be made for both sides of the recruitment process: candidates receive job application suggestions~\cite{martinez2018recommendation}, and hiring teams receive talent suggestions.

In both instances, simply providing the machine learning results as a recommendation obscures the decision-making factors, thereby causing the previously mentioned issues. Counterfactual explanations can effectively address these problems in job recommendations due to their already-mentioned properties: they are easy to comprehend, operate at the instance level, can be applied to any prediction model, and permit customization by allowing the assignment of weights to features. 

\begin{figure}[h]
    \includegraphics[width=\textwidth]{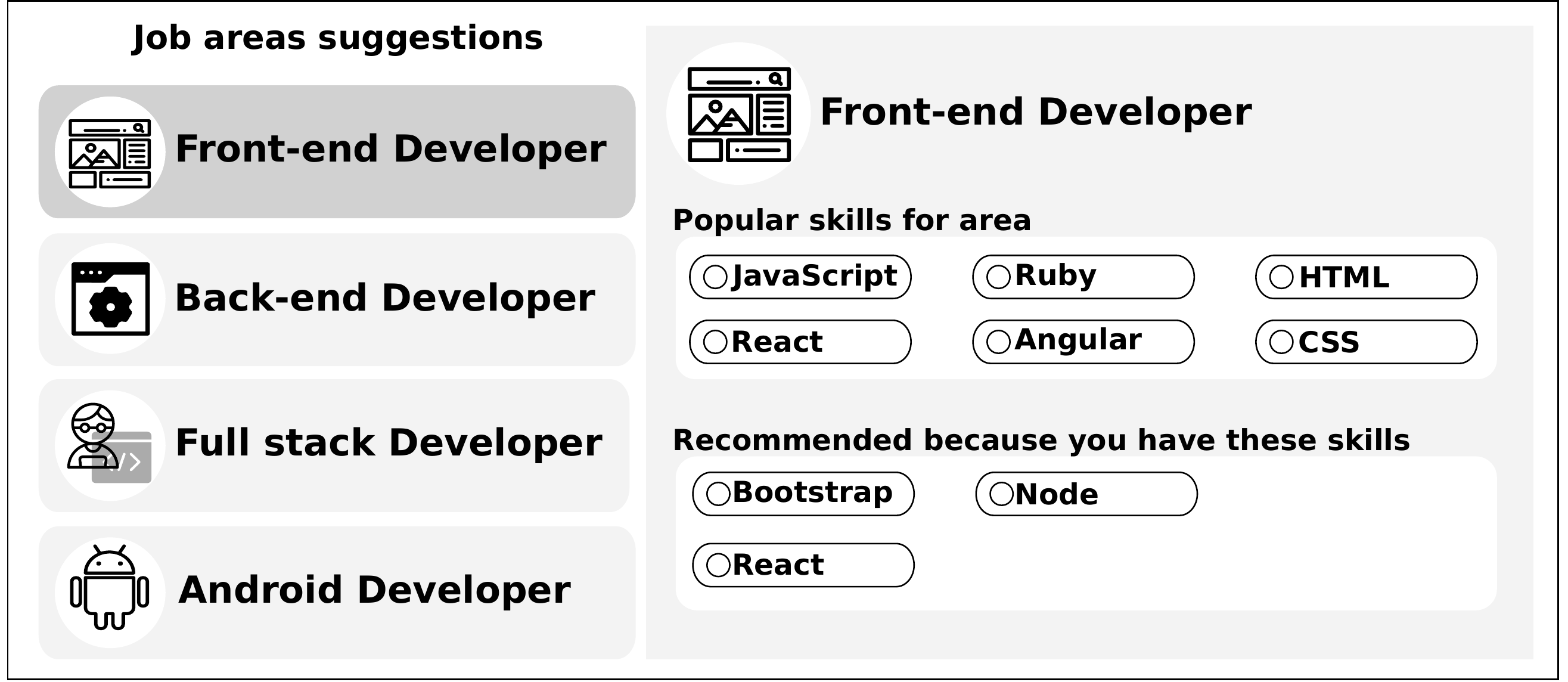}
    \caption{Illustrative example of how a job area suggestion could be explained to the user.}
    \label{fig:applied_cf_cf_suggestion_example}
\end{figure}

Therefore, in the employability context, we can use counterfactual explanations to explain the suggestions derived from models that recommend jobs based on a candidate's characteristics, focusing on highlighting the skills that, if absent, would alter the recommendation. This use case is evaluated in our implementation by producing results similar to those depicted in Figure~\ref{fig:applied_cf_cf_suggestion_example}, where the positive classification result for a determined candidate's CV is explained by indicating which skills should be removed to no longer be a fit for the selected job area.

%In a broader context, counterfactual explanations are primarily used as explanation tools for any model's decision. Therefore, whenever complex models need to offer explanations to users as a way to justify decisions, counterfactual explanations are an efficient method given their decision-oriented nature~\cite{fernandez2020explaining}. However, a recent study~\cite{ramon2021understanding} shows that consumers may prefer feature importance methods over counterfactual explanations in some settings. Interestingly, this study shows that consumers can opt for counterfactual explanations when explaining a negative outcome, where this may correlate to the psychological effects they have on learning new information to avoid future unwanted outcomes~\cite{smallman2012learning}. Nonetheless, we can plot the feature importance methods as bar charts, where we synthesize more information (features and relative importance) in a smaller space, while counterfactual explanations have a textual representation without charts. Therefore, a better, more informative display of counterfactual explanations, combined with its better higher sparsity, could potentially increase its preference by users. Future research can also investigate whether the explanation method has differences depending on the target of explanations (layperson, specialists).

\subsection{Counterfactual as Decision Support}
The use of artificial intelligence raises an important question: how can people, in general, have faith in complex models? This concern is pertinent and crucial to any application in the employability field, as it is a fundamental matter that affects most people's lives. Furthermore, trust is an essential factor in the effectiveness of AI systems in society~\cite{glikson2020human}. Therefore, adopting XAI methodologies can have a positive effect, as explanations can aid in the acceptance of the model. Moreover, counterfactual explanations can be highlighted as an effective method of addressing this challenge because they are based on a thought process commonly used by humans to explain outcomes~\cite{mceleney2006spontaneous} and are therefore intuitively simple to comprehend.

Overall, the interpretability perks of counterfactual explanations may contribute to not just explaining the decision to users but also making them more confident about the model. The literature shows that this improvement in the acceptability of machine learning algorithms has a decisive impact on their incorporation and can foster hypotheses about causality, therefore leading to a better comprehension of the problem~\cite{bidkar2021literature}. However, increased trustworthiness can negatively affect the decision interpretation. This drawback occurs because justifications can induce users to give too much trust in models which are not wholly correct~\cite{ghassemi2021false}. Therefore, prudence and critical thinking may be essential when explanations are given since they are not the root causes for the outcome of the actual problem but rather mathematical inferences obtained from a model. Future research can better explore the handling of trust while maintaining an inquisitive posture towards decisions, where the human-computer interface plays a definitive role. Despite the current challenges, the model's trust is still essential since its absence can critically endanger its use, particularly for high-stake decision fields like employability.

\subsection{Counterfactual as Legal Compliance Method}
Given the significant impact that AI has on society, a number of global initiatives~\cite{commision2021europe,Goodman_Flaxman_2017} aims to develop policies and regulations. Even the EU Commission explicitly identifies AI systems used in the workplace as high-risk, which means that they will be subject to stringent regulations~\cite{commision2021europe}. Particularly significant is the EU General Data Protection Regulation, which mandates that automated decision-making systems provide explanations for their decisions if they can directly affect individuals (which is definitely the case with employability-related applications). In this case, counterfactual explanations have the characteristics required to comply with such regulations~\cite{wachter2017counterfactual}, and they also have the advantage of not disclosing, if correctly treated, the inner workings of the model when explaining decisions~\cite{barocas2020hidden}. This characteristic enhances the security of models that are considered trade secrets.

\subsection{Counterfactual as Guidance}
As previously stated, insights on improving one's employment prospects have significant personal, social, and economic value. Since machine learning models are frequently used for employment prediction~\cite{casuat2019predicting}, XAI methodologies can be utilized (without modifying the original model) to better comprehend why a person has a low probability of entering the labor market.

\begin{figure}[h]
    \includegraphics[width=\textwidth]{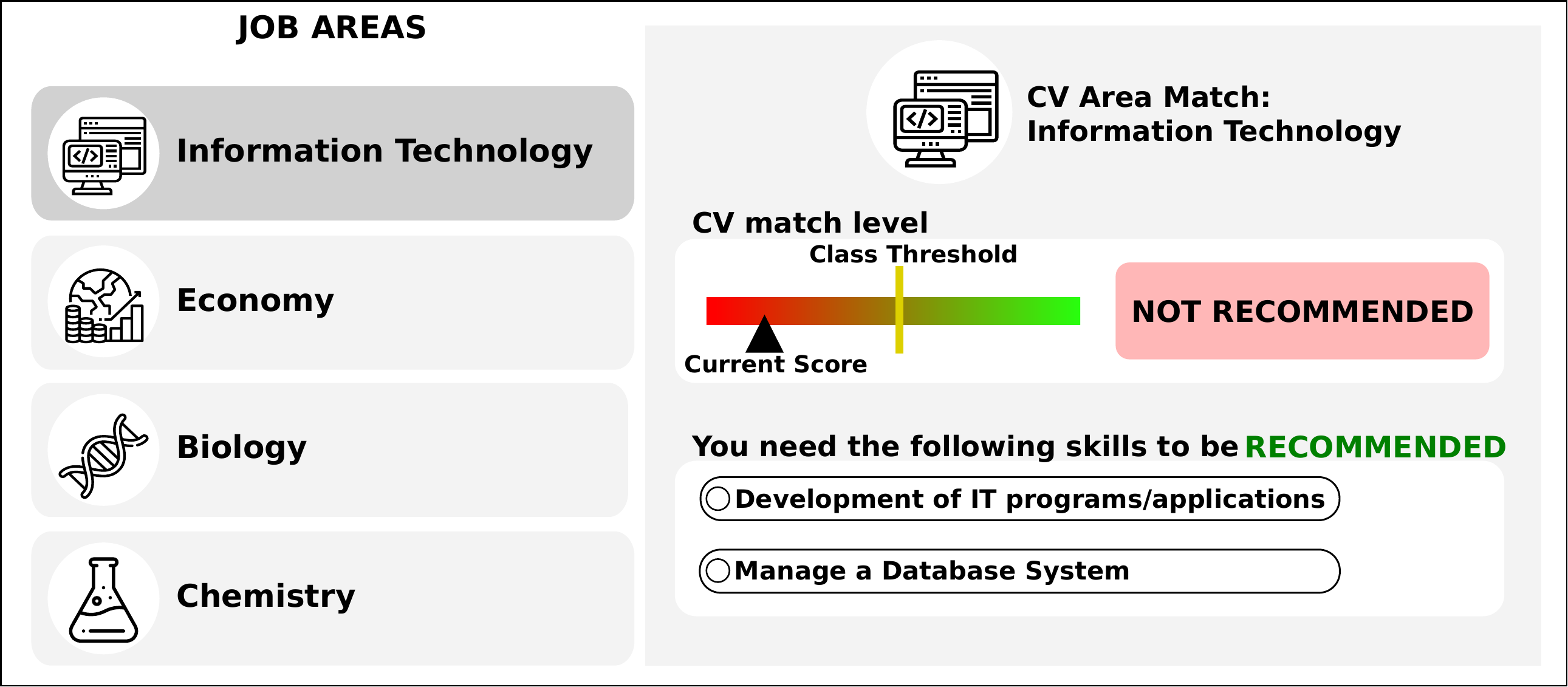}
    \caption{Example of how automated career advice could present to the user, for a specific profession, which skills she/he must acquire to be suitable or more competitive.}
    \label{fig:applied_cf_use_case_cv_match}
\end{figure}

Specifically, counterfactual explanations are well-suited, by definition, to provide guidance regarding what a person must change to achieve a different outcome~\cite{karimi2020algorithmic,barocas2020hidden}.
As a result, they can be used to provide personalized career advice to job-seeking individuals, as the generated explanations highlight the skills that the candidate must acquire to improve their chances of employment. We can use as an illustrative example a simple predictive model that classifies a CV as recommended or not recommended for a particular job field, as demonstrated in Figure~\ref{fig:applied_cf_use_case_cv_match}. In this instance, a counterfactual algorithm will suggest modifying the model's input (the individual's resume skills) to change the predictive class to be recommended. The main difference with the use case of explanations described in Section~\ref{subsec:exp}, is that here we will focus on absent features that, if added, would alter the predicted class, while explanations highlight the present features that, if deleted, would change the predicted class.

This case shows that counterfactuals can have other objectives for the user other than only offering explanations and enhancing trust. Here, we describe its function as guidance to changing the classification output, which consequently includes new constraints. These additional limitations happen because simple or realistic counterfactuals~\cite{de2021framework} are insufficient to give guidance. For the former (simple), it is easy to comprehend since it can suggest unrealistic states, like being married and single at the same time, while for the latter (realistic), although some states are possible, like having 20 year old, this is unachievable to somebody older than that age. Therefore, in this case, counterfactuals must be actionable~\cite{poyiadzi2020face}, which is a characteristic that considers what feasible changes the factual point can make.

\begin{figure}[h]
    \centering
    \includegraphics[width=5cm]{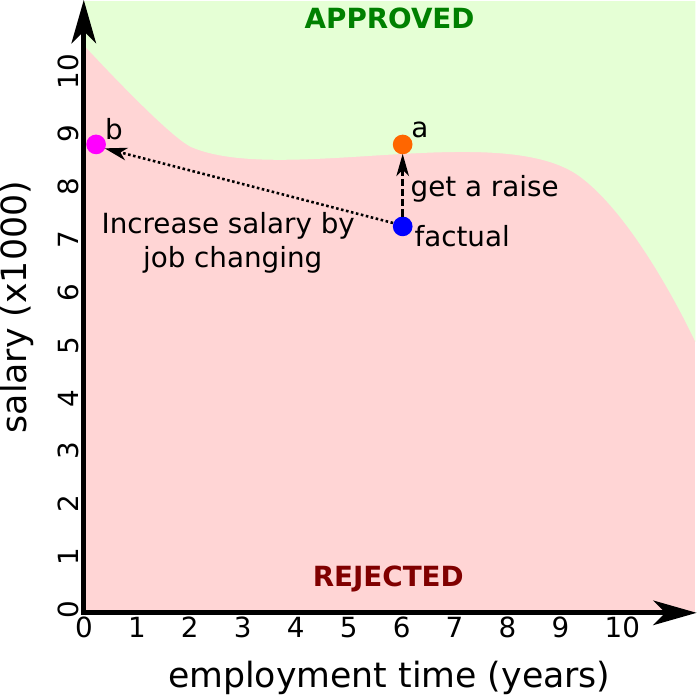}
    \caption{Example of how a counterfactual explanation, \textit{increasing the salary from 7,000 to 9,000}, can lead to different outcomes depending on the action taken. If the person achieves a higher salary by getting a raise (point a), the model will change the classification from rejected to approved, while if the person increases the salary by getting a new job (point b), the change in the classification is not achieved. This happens because the counterfactual considers all other features are constant, therefore, possibly leading to ineffective advice if the person is not aware.}
    \label{fig:applied_cf_example_change_salary}
\end{figure}

Actionable counterfactuals can potentially require drawing a causal relationship between features~\cite{karimi2020algorithmic}. For example, a counterfactual suggesting a single modification, getting a degree, may consider that all other features remain the same, which may not be an actionable change since variables like age are directly related to acquiring new education. Given this causal relationship between features, counterfactuals can even be deceptive~\cite{prosperi2020causal} since the collateral changes provoked by following the counterfactual suggestion may affect the counterfactual objective to change the original class, we depict this scenario in Figure~\ref{fig:applied_cf_example_change_salary}.

Currently, some methods allow defining actionability constraints by manually assigning weights to changes~\cite{brughmans2023nice}, making it easier or more difficult to alter particular characteristics or even create rules that drive feature modifications~\cite{SYNAS}. However, given a large number of features and interactions of some problems, the manual definition of changes can be exhaustive or even impractical. Therefore some methods try to emulate actionability by using statistical or machine learning approaches. For example, FACE~\cite{poyiadzi2020face} uses the dataset to calculate feasible paths by using density metrics, and ALIBI~\cite{ICEGP} uses encoders trained over the original dataset to create a loss term related to counterfactual feasibility. On top of that, we must not discard the role of personal preferences over counterfactual suggestions, where multiple diverse counterfactual explanations, such as done in DiCE~\cite{DICE}, or the interative generation process, can give more autonomy to the user.

In conclusion, for our specific use case applied to employability, we see that counterfactuals can take advantage of complex predictive models to give valuable suggestions that consider both personal skills and market demands. But we can quickly expand this applicability to more general cases where guidance to change the machine learning prediction has some value. In the medical field, for example, this has enormous potential to improve people's lives, although the already cited causal relationship between features and confounding variables represents a fundamental challenge~\cite{prosperi2020causal}. Therefore, future advances in counterfactual generators may expand the functionality of predictive models, promoting them as trustworthy sources of guidance to change undesirable decisions.

\subsection{Counterfactual as Analytical Tool}
\begin{figure}[h]
    \includegraphics[width=10cm]{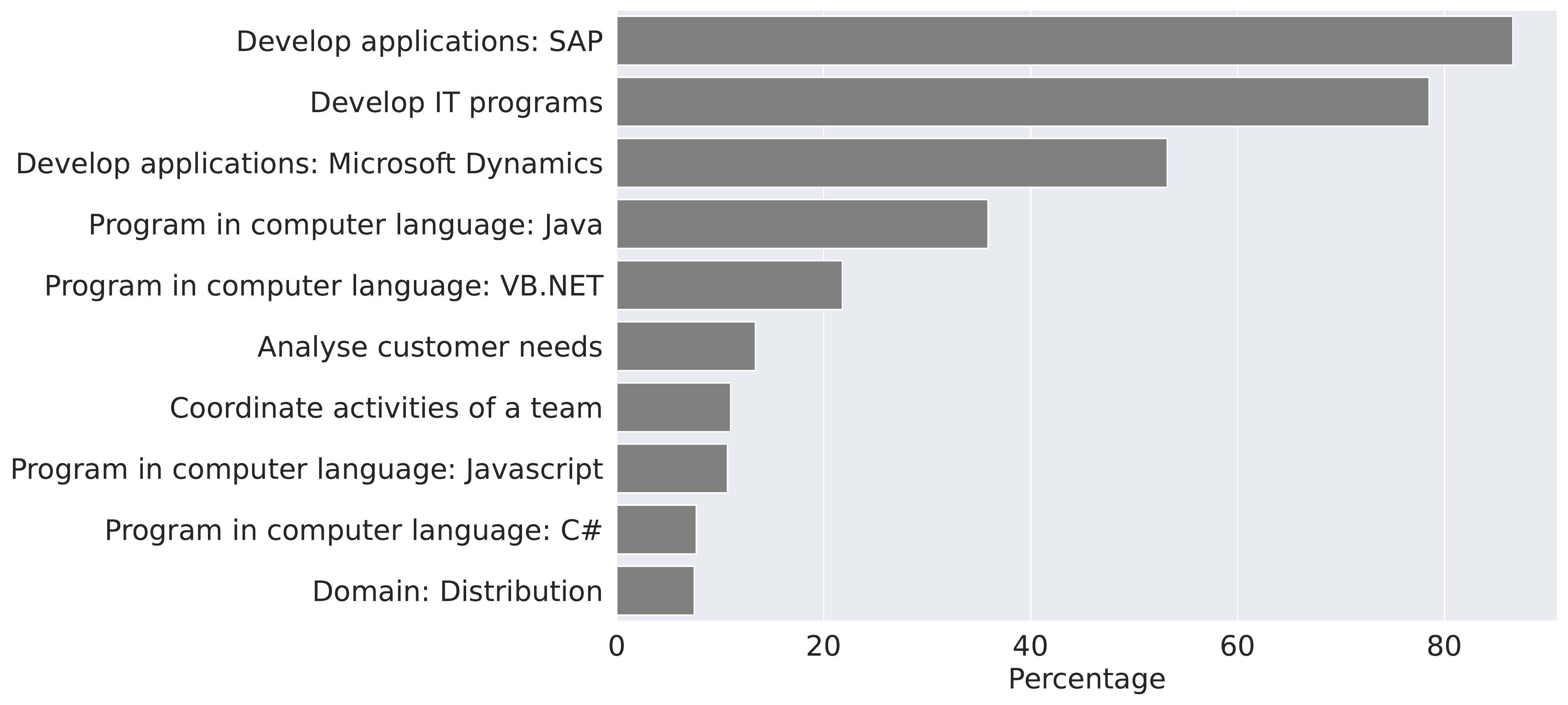}
    \caption{Most frequently occurring missing competences, retrieved from counterfactual explanations derived from decisions made by the predictive model based on people that have a degree related to IT.}
    \label{fig:applied_cf_comp_cf_ai}
\end{figure}

Even though counterfactual explanations provide results for a single instance CV, the aggregation of multiple counterfactual explanations can provide valuable insights into the modeled problem. Figure~\ref{fig:applied_cf_comp_cf_ai} depicts the most cited competencies in numerous counterfactual explanations generated over factual CVs for the IT field; therefore, this chart identifies which skills, if obtained, will contribute the most to changing the classification of the candidate to a high job reach. This information is valuable because it considers both sides of employment, the skills of candidates, and the needs of employers, thereby optimizing the skill match.

\begin{figure}[h]
    \includegraphics[width=10cm]{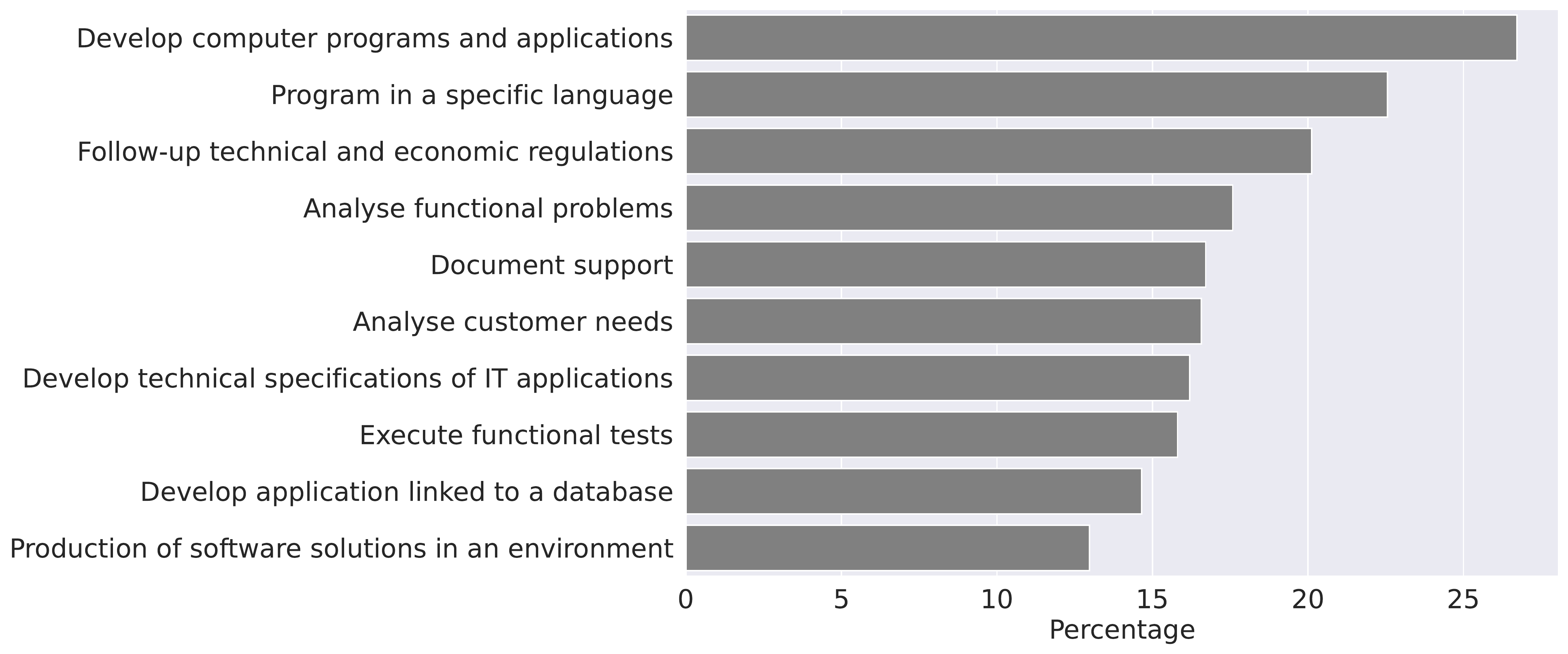}
    \caption{Most requested competencies asked in IT related positions.}
    \label{fig:applied_cf_overall_job_ai}
\end{figure}

Therefore, multiple counterfactuals can provide a new view where the values represent modifications to alter the prediction of the analyzed instances. We can highlight the utility of this approach if we consider the following theoretical case: a governmental institute wants to know which skills they should offer to improve employability levels. The most obvious approach is to find the skills most requested by employers; we did this with VDAB data in Figure~\ref{fig:applied_cf_overall_job_ai}. However, if we consider people learning the two most requested skills, the employability enhancement, measured by the people classified as having high job reach, is worse than the top suggestions made by counterfactuals. We depict this result in Table~\ref{tab:vdab_ai_avg_cf_comparison_transposed}, which shows the superiority of counterfactual explanations in finding the most effective set of features to change the prediction outcome.

We then can expand this use case to any application where the model's decision is a relevant metric to be enhanced in a population. The characteristic of considering individual and dataset aspects makes aggregate counterfactual explanations a potentially helpful analysis to find which minimal set of changes can lead to the best impact on the objective score, especially in high non-linear models in which individual feature contributions may not consistently affect decisions~\cite{fernandez2020explaining}. In addition, the aggregated review of the explanations can create new knowledge. For example, in the medical field, studies~\cite{arcadu2019deep} detected relevant parameters that affected the decision of complex machine learning models by evaluating the explanation of various instances.

\begin{table}[ht]
\centering
\begin{tabular}{@{}p{3cm}*{10}{p{0.75cm}}@{}}
\toprule
& 1 & 2 & 3 & 4 & 5 & 6 & 7 & 8 & 9 & 10 \\ 
\midrule
\textbf{Top Selected} & & & & & & & & & & \\
\quad AVG (\%) & 1.6 & 3.5 & 3.2 & 5.1 & 5.9 & 11.8 & 17.6 & 26.4 & 29.6 & 48.0 \\
\quad SEDC (\%) & 9.1 & 15.3 & 34.5 & 70.5 & 78.4 & 89.9 & 92.8 & 97.1 & 98.2 & 98.3 \\ 
\bottomrule
\end{tabular}
\caption{Table showing the percentage of instances with classification modified to a favorable outcome when changing 1 to 10 top features of most requested competencies asked by IT positions (AVG) and the top 10 most frequently counterfactual features.}
\label{tab:vdab_ai_avg_cf_comparison_transposed}
\end{table}

\section{Additional Use Cases}

The last section used the VDAB data to exemplify five different use cases in which counterfactuals can be used, including its classical explanation case and additional cases related to decision support, compliance, guidance, and analytics. However, although not tested with the data, we identified three additional use cases in which counterfactuals could assist in the employability field. The first one is related to bias detection. Although counterfactuals cannot solve the problem of biased models, they can be a helpful tool to evaluate if protected variables change the prediction decisions. This application can consist remarkable repercussions on employment-related tasks since historical data on which machine learning algorithms are trained may include historical bias against minority groups, for example.

The second use case is related to the applicability of the counterfactual results to debug possible unexpected behaviors leading to misclassifications. Given the decision nature of counterfactual explanations, it can show the features that would revert such behavior and, consequently, lead to improvements or a better understanding of the underlying causes for the model's behavior. The third use case is also related to a technical aspect, where the possibility of automatically generating counterfactuals by using the multiple generative methods (present in literature) allows the adaptation of it as a pipe segment in an MLOps framework, which specific business rules could be checked (such as a bias for certain features) and controlled.

\section{Counterfactual Explanation Limitations}

Despite the multiple application cases in which counterfactual explanations can be useful, there are also circumstances that they are not optimal or even advisable. The most significant inappropriate application is when prediction scoring, rather than prediction decision, is the subject of study. Therefore, counterfactual explanations do not give a clear picture of how the variation of features influences the prediction score. For this case, popular feature importance methods such as LIME~\cite{ribeiro2016should} and SHAP~\cite{lundberg2017unified} are much more appropriate since they evaluate how each feature influences the model's prediction scoring. Moreover, counterfactual explanations can represent a threat to intellectual property if not correctly used, as additional points of different classes can be used in model extraction attacks~\cite{aivodji2020model}, which can create faithful copies of the original model even under low query budgets~\cite{aivodji2020model}. Multiple counterfactuals can also empower malicious users to play with the model, allowing them to exploit possible flaws~\cite{pawelczyk2022exploring}, similar to what is done with adversarial attacks. If not properly treated, counterfactual explanations also create privacy concerns as the feature changes and background information can lead to the so-called \textit{explanation linkage attack}~\cite{goethals2022privacy}, which may possibly identify anonymized individuals. Finally, counterfactual explanations can be misleading in several conditions. For example, as we previously discussed, if the model is not reliable enough,  counterfactual changes may not reflect a change in the outcome in reality. Also, if the model changes over time (a typical occurrence in real-world production environments), a counterfactual may not be valid in different versions and then be a source of confusion or dispute.
For the employment context, all these limitations are relevant since they can be applicable to any kind of model and problem. Therefore, practitioners must carefully evaluate those concerns and, if one or more are pertinent, find possible solutions or avoid using them. %We can illustrate one of these concerns by considering a theoretical case in which counterfactual explanations use some personal information for job fit assessments. As mentioned above, \textit{explanation linkage attacks}~\cite{goethals2022privacy} can be performed to retrieve private information from other people that should be anonymous. For this theoretical case, a possible solution could include methods such as anonymization of data or counterfactual anonymization treatments~\cite{goethals2022privacy}. Therefore, this also shows that more research is needed to evaluate the potential risks associated with the usage of counterfactual explanations.

\section{Conclusion}
The advancements in artificial intelligence have led to the creation of powerful methodologies that have altered the manner in which we analyze complex problems, such as those pertaining to employability. These highly efficient methodologies are notable for their complexity, which contributes to their inherent inexplicability. As the EU Commission identifies the employment sector as a high-risk area for the use of AI~\cite{commision2021europe}, it is essential to provide sufficient information on these methodologies. The XAI field provides the means to address this issue: we demonstrated how a specific XAI methodology, counterfactual explanations, can be applied in the employability field by presenting five real use cases that result in more interpretable decisions and beyond. These use cases reach diverse objectives and stakeholders: career advice for potential hires, explanations of job recommendations to job posting systems applicants, gaining new insights from models and data for institutions and policymakers, gaining trust and social acceptance for the general public, and legal compliance for authorities. We anticipate that this novel perspective on the application of counterfactual explanations will provide the necessary insights to resolve current issues and inspire future research in the field. In addition, we anticipate that future work can include other novel objectives such as bias detection and model improvement. Finally, since there is a great degree of variation in how counterfactuals can be generated~\cite{de2021framework} (where multiple implementation strategies are described in the literature), prospective researchers can also examine how these differences affect results for particular use cases.

\section*{Acknowledgements}
This research received funding from the Flemish Government (AI Research Program) and used secondary, pseudonimized HR data provided by VDAB.

\printbibliography %Prints bibliography

@inproceedings{ribeiro2016should,
  title={" Why should i trust you?" Explaining the predictions of any classifier},
  author={Ribeiro, Marco Tulio and Singh, Sameer and Guestrin, Carlos},
  booktitle={Proceedings of the 22nd ACM SIGKDD international conference on knowledge discovery and data mining},
  pages={1135--1144},
  year={2016}
}

@article{lundberg2017unified,
  title={A unified approach to interpreting model predictions},
  author={Lundberg, Scott M and Lee, Su-In},
  journal={Advances in neural information processing systems},
  volume={30},
  year={2017}
}

@article{fernandez2020explaining,
  title={Explaining data-driven decisions made by AI systems: The counterfactual approach},
  author={Fernandez-Loria, Carlos and Provost, Foster and Han, Xintian},
  journal={arXiv preprint arXiv:2001.07417},
  year={2020}
}

@article{martens2014explaining,
  title={Explaining data-driven document classifications},
  author={Martens, David and Provost, Foster},
  journal={MIS quarterly},
  volume={38},
  number={1},
  pages={73--100},
  year={2014},
  publisher={JSTOR}
}

@article{de2021framework,
  title={A framework and benchmarking study for counterfactual generating methods on tabular data},
  author={de Oliveira, Raphael Mazzine Barbosa and Martens, David},
  journal={Applied Sciences},
  volume={11},
  number={16},
  pages={7274},
  year={2021},
  publisher={MDPI}
}

@article{karimi2020algorithmic,
  title={Algorithmic Recourse: from Counterfactual Explanations to Interventions},
  author={Karimi, Amir-Hossein and Scholkopf, Bernhard and Valera, Isabel},
  journal={arXiv preprint arXiv:2002.06278},
  year={2020}
}

@article{DICE,
  author    = {Ramaravind Kommiya Mothilal and
               Amit Sharma and
               Chenhao Tan},
  title     = {Explaining Machine Learning Classifiers through Diverse Counterfactual
               Explanations},
  journal   = {CoRR},
  volume    = {abs/1905.07697},
  year      = {2019},
  url       = {http://arxiv.org/abs/1905.07697},
  archivePrefix = {arXiv},
  eprint    = {1905.07697},
  bibsource = {dblp computer science bibliography, https://dblp.org}
}

@article{SYNAS,
  title={Synthesizing Action Sequences for Modifying Model Decisions},
  author={Ramakrishnan, Goutham and Lee, Yun Chan and Albarghouthi, Aws},
  journal={arXiv preprint arXiv:1910.00057},
  year={2019}
}

@article{ICEGP,
  author    = {Arnaud Van Looveren and
               Janis Klaise},
  title     = {Interpretable Counterfactual Explanations Guided by Prototypes},
  journal   = {CoRR},
  volume    = {abs/1907.02584},
  year      = {2019},
  url       = {http://arxiv.org/abs/1907.02584},
  archivePrefix = {arXiv},
  eprint    = {1907.02584},
  bibsource = {dblp computer science bibliography, https://dblp.org}
}

@article{miller2019explanation,
  title={Explanation in artificial intelligence: Insights from the social sciences},
  author={Miller, Tim},
  journal={Artificial Intelligence},
  volume={267},
  pages={1--38},
  year={2019},
  publisher={Elsevier}
}

@article{wachter2017counterfactual,
  title={Counterfactual explanations without opening the black box: Automated decisions and the GDPR},
  author={Wachter, Sandra and Mittelstadt, Brent and Russell, Chris},
  journal={Harv. JL \& Tech.},
  volume={31},
  pages={841},
  year={2017},
  publisher={HeinOnline}
}

@article{hillage1998employability,
  title={Employability: developing a framework for policy analysis},
  author={Hillage, Jim and Pollard, Emma},
  year={1998},
  publisher={DfEE London},
  journal={DfEE Publications}
}

@article{pheko2017addressing,
  title={Addressing employability challenges: a framework for improving the employability of graduates in Botswana},
  author={Pheko, Mpho M and Molefhe, Kaelo},
  journal={International Journal of Adolescence and Youth},
  volume={22},
  number={4},
  pages={455--469},
  year={2017},
  publisher={Taylor \& Francis}
}

@inproceedings{klosters2014matching,
  title={Matching skills and labour market needs: Building social partnerships for better skills and better jobs},
  author={Klosters, Davos},
  booktitle={World Economic Forum Global Agenda Council on Employment},
  pages={1--28},
  year={2014}
}

@article{jantawan2013application,
  title={The application of data mining to build classification model for predicting graduate employment},
  author={Jantawan, Bangsuk and Tsai, Cheng-Fa},
  journal={arXiv preprint arXiv:1312.7123},
  year={2013}
}

@article{boselli2018classifying,
  title={Classifying online job advertisements through machine learning},
  author={Boselli, Roberto and Cesarini, Mirko and Mercorio, Fabio and Mezzanzanica, Mario},
  journal={Future Generation Computer Systems},
  volume={86},
  pages={319--328},
  year={2018},
  publisher={Elsevier}
}

@inproceedings{mujtaba2019ethical,
  title={Ethical considerations in ai-based recruitment},
  author={Mujtaba, Dena F and Mahapatra, Nihar R},
  booktitle={2019 IEEE International Symposium on Technology and Society (ISTAS)},
  pages={1--7},
  year={2019},
  organization={IEEE}
}

@inproceedings{martinez2018recommendation,
  title={Recommendation of job offers using random forests and support vector machines},
  author={Martinez-Gil, Jorge and Freudenthaler, Bernhard and Natschlager, Thomas},
  booktitle={Proceedings of the Workshops of the EDBT/ICDT 2018 Joint Conference (EDBT/ICDT 2018)},
  pages={22--27},
  year={2018}
}

@inproceedings{ochmann2021evaluation,
  title={The evaluation of the black box problem for AI-based recommendations: An interview-based study},
  author={Ochmann, Jessica and Zilker, Sandra and Laumer, Sven},
  booktitle={Proceedings of the 16. Internationale Tagung Wirtschaftsinformatik},
  year={2021},
  pages={1--16}
}

@article{goethals2022comprehensibility,
  title={The non-linear nature of the cost of comprehensibility},
  author={Goethals, Sofie and Martens, David and Evgeniou, Theodoros},
  journal={Journal of Big Data},
  volume={9},
  number={1},
  pages={1--23},
  year={2022},
  publisher={Springer}
}

@article{rai2020explainable,
  title={Explainable AI: From black box to glass box},
  author={Rai, Arun},
  journal={Journal of the Academy of Marketing Science},
  volume={48},
  number={1},
  pages={137--141},
  year={2020},
  publisher={Springer}
}

@inproceedings{sokol2019counterfactual,
  title={Counterfactual explanations of machine learning predictions: opportunities and challenges for AI safety},
  author={Sokol, Kacper and Flach, Peter A},
  booktitle={SafeAI@ AAAI},
  pages={1--4},
  year={2019}
}

@inproceedings{mashayekhi2021quantifying,
  title={Quantifying and reducing imbalance in networks},
  author={Mashayekhi, Yoosof and Kang, Bo and Lijffijt, Jefrey and De Bie, Tijl},
  booktitle={RECSYS IN HR 2021},
  volume={2967},
  year={2021},
  pages={1--10},
  organization={CEUR}
}

@article{ribeiro2016model,
  title={Model-agnostic interpretability of machine learning},
  author={Ribeiro, Marco Tulio and Singh, Sameer and Guestrin, Carlos},
  journal={arXiv preprint arXiv:1606.05386},
  year={2016}
}

@article{lundberg2020local,
  title={From local explanations to global understanding with explainable AI for trees},
  author={Lundberg, Scott M and Erion, Gabriel and Chen, Hugh and DeGrave, Alex and Prutkin, Jordan M and Nair, Bala and Katz, Ronit and Himmelfarb, Jonathan and Bansal, Nisha and Lee, Su-In},
  journal={Nature machine intelligence},
  volume={2},
  number={1},
  pages={56--67},
  year={2020},
  publisher={Nature Publishing Group}
}

@article{ramon2020comparison,
  title={A comparison of instance-level counterfactual explanation algorithms for behavioral and textual data: SEDC, LIME-C and SHAP-C},
  author={Ramon, Yanou and Martens, David and Provost, Foster and Evgeniou, Theodoros},
  journal={Advances in Data Analysis and Classification},
  volume={14},
  number={4},
  pages={801--819},
  year={2020},
  publisher={Springer}
}

@article{spooner2021counterfactual,
  title={Counterfactual Explanations for Arbitrary Regression Models},
  author={Spooner, Thomas and Dervovic, Danial and Long, Jason and Shepard, Jon and Chen, Jiahao and Magazzeni, Daniele},
  journal={arXiv preprint arXiv:2106.15212},
  year={2021}
}

@inproceedings{hada2021exploring,
  title={Exploring counterfactual explanations for classification and regression trees},
  author={Hada, Suryabhan Singh and Carreira-Perpinan, Miguel A},
  booktitle={Joint European Conference on Machine Learning and Knowledge Discovery in Databases},
  pages={489--504},
  year={2021},
  organization={Springer}
}

@article{alfano2020technologically,
  title={Technologically scaffolded atypical cognition: the case of YouTube’s recommender system},
  author={Alfano, Mark and Fard, Amir Ebrahimi and Carter, J Adam and Clutton, Peter and Klein, Colin},
  journal={Synthese},
  pages={1--24},
  year={2020},
  publisher={Springer}
}

@inproceedings{casuat2019predicting,
  title={Predicting students' employability using machine learning approach},
  author={Casuat, Cherry D and Festijo, Enrique D},
  booktitle={2019 IEEE 6th International Conference on Engineering Technologies and Applied Sciences (ICETAS)},
  pages={1--5},
  year={2019},
  organization={IEEE}
}

@inproceedings{barocas2020hidden,
  title={The hidden assumptions behind counterfactual explanations and principal reasons},
  author={Barocas, Solon and Selbst, Andrew D and Raghavan, Manish},
  booktitle={Proceedings of the 2020 Conference on Fairness, Accountability, and Transparency},
  pages={80--89},
  year={2020}
}

@article{brughmans2023nice,
  title={Nice: an algorithm for nearest instance counterfactual explanations},
  author={Brughmans, Dieter and Leyman, Pieter and Martens, David},
  journal={Data Mining and Knowledge Discovery},
  pages={1--39},
  year={2023},
  publisher={Springer}
}

@inproceedings{poyiadzi2020face,
  title={FACE: feasible and actionable counterfactual explanations},
  author={Poyiadzi, Rafael and Sokol, Kacper and Santos-Rodriguez, Raul and De Bie, Tijl and Flach, Peter},
  booktitle={Proceedings of the AAAI/ACM Conference on AI, Ethics, and Society},
  pages={344--350},
  year={2020}
}

@article{prosperi2020causal,
  title={Causal inference and counterfactual prediction in machine learning for actionable healthcare},
  author={Prosperi, Mattia and Guo, Yi and Sperrin, Matt and Koopman, James S and Min, Jae S and He, Xing and Rich, Shannan and Wang, Mo and Buchan, Iain E and Bian, Jiang},
  journal={Nature Machine Intelligence},
  volume={2},
  number={7},
  pages={369--375},
  year={2020},
  publisher={Nature Publishing Group}
}

@article{arcadu2019deep,
  title={Deep learning algorithm predicts diabetic retinopathy progression in individual patients},
  author={Arcadu, Filippo and Benmansour, Fethallah and Maunz, Andreas and Willis, Jeff and Haskova, Zdenka and Prunotto, Marco},
  journal={NPJ digital medicine},
  volume={2},
  number={1},
  pages={1--9},
  year={2019},
  publisher={Nature Publishing Group}
}

@article{glikson2020human,
  title={Human trust in artificial intelligence: Review of empirical research},
  author={Glikson, Ella and Woolley, Anita Williams},
  journal={Academy of Management Annals},
  volume={14},
  number={2},
  pages={627--660},
  year={2020},
  publisher={Briarcliff Manor, NY}
}

@article{mceleney2006spontaneous,
  title={Spontaneous counterfactual thoughts and causal explanations},
  author={McEleney, Alice and Byrne, Ruth MJ},
  journal={Thinking \& Reasoning},
  volume={12},
  number={2},
  pages={235--255},
  year={2006},
  publisher={Taylor \& Francis}
}

@article{bidkar2021literature,
  title={A literature review to underline necessity of explainability in AI and discuss existing explainable AI techniques},
  author={Bidkar, Deepti Vinayak},
  year={2021}
}

@article{ghassemi2021false,
  title={The false hope of current approaches to explainable artificial intelligence in health care},
  author={Ghassemi, Marzyeh and Oakden-Rayner, Luke and Beam, Andrew L},
  journal={The Lancet Digital Health},
  volume={3},
  number={11},
  pages={e745--e750},
  year={2021},
  publisher={Elsevier}
}

@article{Goodman_Flaxman_2017, title={European Union Regulations on Algorithmic Decision-Making and a “Right to Explanation”}, volume={38}, url={https://ojs.aaai.org/index.php/aimagazine/article/view/2741}, DOI={10.1609/aimag.v38i3.2741}, abstractNote={We summarize the potential impact that the European Union’s new General Data Protection Regulation will have on the routine use of machine learning algorithms. Slated to take effect as law across the EU in 2018, it will restrict automated individual decision-making (that is, algorithms that make decisions based on user-level predictors) which “significantly affect” users. The law will also effectively create a “right to explanation,” whereby a user can ask for an explanation of an algorithmic decision that was made about them. We argue that while this law will pose large challenges for industry, it highlights opportunities for computer scientists to take the lead in designing algorithms and evaluation frameworks which avoid discrimination and enable explanation.}, number={3}, journal={AI Magazine}, author={Goodman, Bryce and Flaxman, Seth}, year={2017}, month={Oct.}, pages={50-57} }

@article{commision2021europe,
  title={Europe Fit for the Digital Age: Commission Proposes New Rules and Actions for Excellence and Trust in Artificial Intelligence},
  author={Commision, Europan},
  journal={Europan Commision: Geneva, Switzerland},
  year={2021}
}

@article{aivodji2020model,
  title={Model extraction from counterfactual explanations},
  author={Aivodji, Ulrich and Bolot, Alexandre and Gambs, S{\'e}bastien},
  journal={arXiv preprint arXiv:2009.01884},
  year={2020}
}

@inproceedings{pawelczyk2022exploring,
  title={Exploring counterfactual explanations through the lens of adversarial examples: A theoretical and empirical analysis},
  author={Pawelczyk, Martin and Agarwal, Chirag and Joshi, Shalmali and Upadhyay, Sohini and Lakkaraju, Himabindu},
  booktitle={International Conference on Artificial Intelligence and Statistics},
  pages={4574--4594},
  year={2022},
  organization={PMLR}
}

@article{goethals2022privacy,
  title={The privacy issue of counterfactual explanations: explanation linkage attacks},
  author={Goethals, Sofie and S{\"o}rensen, Kenneth and Martens, David},
  journal={arXiv preprint arXiv:2210.12051},
  year={2022}
}
\end{document}